\documentclass[11pt]{article}

\usepackage[a4paper,margin=2.5cm,heightrounded=true]{geometry}
\usepackage{times}
\usepackage{latexsym}
\usepackage[T1]{fontenc}
\usepackage[utf8]{inputenc}
\usepackage{microtype}
\usepackage{inconsolata}
\usepackage{graphicx}
\usepackage{booktabs}
\usepackage{array}
\usepackage{amsmath,amssymb}
\usepackage{mathtools}
\usepackage{natbib}
\usepackage{xcolor}
\usepackage[switch,mathlines]{lineno}
\usepackage{etoolbox}
\usepackage{caption}
\usepackage{url}
\usepackage[breaklinks]{hyperref}
\usepackage{placeins}
\usepackage{balance}
\usepackage{xspace}
\usepackage{multirow}
\usepackage{enumitem}
\setlist{nosep,leftmargin=*}

\twocolumn
\makeatletter

\newcount\cv@tmpc@ \newcount\cv@tmpc
\def\fillzeros[#1]#2{\cv@tmpc@=#2\relax\ifnum\cv@tmpc@<0\cv@tmpc@=-\cv@tmpc@\fi
  \cv@tmpc=1 %
  \loop\ifnum\cv@tmpc@<10 \else \divide\cv@tmpc@ by 10 \advance\cv@tmpc by 1 \fi
    \ifnum\cv@tmpc@=10\relax\cv@tmpc@=11\relax\fi \ifnum\cv@tmpc@>10 \repeat
  \ifnum#2<0\advance\cv@tmpc1\relax-\fi
  \loop\ifnum\cv@tmpc<#1\relax0\advance\cv@tmpc1\relax\fi \ifnum\cv@tmpc<#1 \repeat
  \cv@tmpc@=#2\relax\ifnum\cv@tmpc@<0\cv@tmpc@=-\cv@tmpc@\fi \relax\the\cv@tmpc@}%

\newcommand*\linenomathpatch[1]{%
  \expandafter\pretocmd\csname #1\endcsname {\linenomath}{}{}%
  \expandafter\pretocmd\csname #1*\endcsname {\linenomath}{}{}%
  \expandafter\apptocmd\csname end#1\endcsname {\endlinenomath}{}{}%
  \expandafter\apptocmd\csname end#1*\endcsname {\endlinenomath}{}{}%
}
\newcommand*\linenomathpatchAMS[1]{%
  \expandafter\pretocmd\csname #1\endcsname {\linenomathAMS}{}{}%
  \expandafter\pretocmd\csname #1*\endcsname {\linenomathAMS}{}{}%
  \expandafter\apptocmd\csname end#1\endcsname {\endlinenomath}{}{}%
  \expandafter\apptocmd\csname end#1*\endcsname {\endlinenomath}{}{}%
}
\expandafter\ifx\linenomath\linenomathWithnumbers
  \let\linenomathAMS\linenomathWithnumbers
  \patchcmd\linenomathAMS{\advance\postdisplaypenalty\linenopenalty}{}{}{}
\else
  \let\linenomathAMS\linenomathNonumbers
\fi
\AtBeginDocument{%
  \linenomathpatch{equation}%
  \linenomathpatchAMS{gather}%
  \linenomathpatchAMS{multline}%
  \linenomathpatchAMS{align}%
  \linenomathpatchAMS{alignat}%
  \linenomathpatchAMS{flalign}%
}

\newlength\titlebox
\renewcommand\maketitle{\par
 \begingroup
   \def\thefootnote{\fnsymbol{footnote}}
   \twocolumn[\@maketitle]
   \@thanks
 \endgroup
 \setcounter{footnote}{0}
 \let\maketitle\relax
 \let\@maketitle\relax
 \gdef\@thanks{}\gdef\@author{}\gdef\@title{}\let\thanks\relax}
\def\@maketitle{\vbox to \titlebox{\hsize\textwidth
 \linewidth\hsize \vskip 0.125in minus 0.125in \centering
 {\Large\bfseries \@title \par} \vskip 0.2in plus 1fil minus 0.1in
 {\large \@author \par}
 \vskip 0.3in plus 2fil minus 0.1in
}}

\renewenvironment{abstract}%
  {\begin{center}\large\textbf{\abstractname}\end{center}%
   \begin{list}{}{%
     \setlength{\rightmargin}{0.6cm}%
     \setlength{\leftmargin}{0.6cm}}%
   \item[]\ignorespaces%
   \@setsize\normalsize{12pt}\xpt\@xpt}%
  {\unskip\end{list}}

\def\section{\@startsection {section}{1}{\z@}{-2.0ex plus -0.5ex minus -.2ex}{1.5ex plus 0.3ex minus .2ex}{\large\bfseries\raggedright}}
\def\subsection{\@startsection{subsection}{2}{\z@}{-1.8ex plus -0.5ex minus -.2ex}{0.8ex plus .2ex}{\normalsize\bfseries\raggedright}}
\def\subsubsection{\@startsection{subsubsection}{3}{\z@}{-1.5ex plus -0.5ex minus -.2ex}{0.5ex plus .2ex}{\normalsize\bfseries\raggedright}}
\def\paragraph{\@startsection{paragraph}{4}{\z@}{1.5ex plus 0.5ex minus .2ex}{-1em}{\normalsize\bfseries}}
\def\subparagraph{\@startsection{subparagraph}{5}{\parindent}{1.5ex plus 0.5ex minus .2ex}{-1em}{\normalsize\bfseries}}

\def\thebibliography#1{\vskip\parskip%
\vskip\baselineskip%
\def\baselinestretch{1}%
\ifx\@currsize\normalsize\@normalsize\else\@currsize\fi%
\vskip-\parskip
\section*{References\@mkboth{References}{References}}\list
 {}{\setlength{\labelwidth}{0pt}\setlength{\leftmargin}{\parindent}%
 \setlength{\itemindent}{-\parindent}}%
 \def\newblock{\hskip .11em plus .33em minus -.07em}%
 \sloppy\clubpenalty4000\widowpenalty4000\sfcode`\.=1000\relax}

\labelwidth\leftmargini\advance\labelwidth-\labelsep \labelsep 5pt
\def\@listi{\leftmargin\leftmargini}
\def\@listii{\leftmargin\leftmarginii
 \labelwidth\leftmarginii\advance\labelwidth-\labelsep
 \topsep 2pt plus 1pt minus 0.5pt
 \parsep 1pt plus 0.5pt minus 0.5pt
 \itemsep \parsep}
\belowdisplayskip \abovedisplayskip
\def\@normalsize{\@setsize\normalsize{11pt}\xpt\@xpt}
\def\small{\@setsize\small{10pt}\ixpt\@ixpt}
\def\footnotesize{\@setsize\footnotesize{10pt}\ixpt\@ixpt}
\def\scriptsize{\@setsize\scriptsize{8pt}\viipt\@viipt}
\def\tiny{\@setsize\tiny{7pt}\vipt\@vipt}
\def\large{\@setsize\large{14pt}\xiipt\@xiipt}
\def\Large{\@setsize\Large{16pt}\xivpt\@xivpt}
\makeatother

\definecolor{darkblue}{rgb}{0,0,0.5}
\hypersetup{colorlinks=true,citecolor=darkblue,linkcolor=darkblue,urlcolor=darkblue}
\hypersetup{pdftitle={AfriSyCo: Measuring Assertive Framing, Verification, and Wording Sensitivity Around African-Language Content},pdfauthor={David Ababio Awuni; Rose-Mary Owusuaa Mensah Gyening; Elvis Gyasi Owusu}}

\newcommand{\pp}{~pp\xspace}

\newcommand{\afrisyco}{\textsc{AfriSyCo}\xspace}

\title{AfriSyCo: Measuring Assertive Framing, Verification, and Wording Sensitivity\\Around African-Language Content}
\author{David Ababio Awuni \quad Rose-Mary Owusuaa Mensah Gyening\thanks{Corresponding author.}\\
Elvis Gyasi Owusu}
\date{}

\begin{document}
\maketitle

\begin{abstract}
\afrisyco studies answer switching around African-language factual content with two complementary layers: native-language follow-ups and a controlled cross-language factorial whose question, options, and target remain in the African language while the follow-up framing is English. We analyze 1,415 turn-1-correct model--language--item observations derived from 100 source questions across seven open-weight checkpoints and six languages; turn-1-correct denotes observed first-response accuracy, not demonstrated knowledge. Under native prompts, assertive endorsement produces 29.3 percentage points more any-turn false-target selection than mention-plus-verification (M+V), with a 19.0-point immediate T2 contrast. In the precommitted $2\times2$ factorial, averaged over three tested prompt families, assertive framing increases target selection by 30.4 points (95\% CI [28.4, 32.3]); verification decreases it by 17.4 points, while the assertive effect rises from 20.5 points without verification to 40.2 with it (interaction +19.7). The effect remains 34.8 points among 611 observations correct after option reordering. Magnitude varies sharply by wording and checkpoint: prompt-family effects span 20.1--42.5 points, a Twi/Qwen3 paraphrase shifts target selection from 70.8\% to 4.2\%, and checkpoint effects span 9.2--47.0 points. Prompt realization is therefore part of the measurement problem.
\end{abstract}

\section{Introduction}

Suppose a model answers a factual multiple-choice question correctly. A user then names another option, states that it is correct, and asks the model to verify. If the model switches, the behavior is easy to label ``sycophancy.'' Yet that single follow-up bundles several cues: the target is mentioned, its correctness is asserted, and verification is requested. A switch rate from such a prompt cannot show which cue moved the model.

This measurement problem matters because sycophancy research spans static agreement, rebuttal, persuasion, multi-turn interaction, authority, and multilingual settings \citep{sharma2024,fanous2025,hong2025,kim2025,tan2025,ranaldi2026,maraia2026}. The construct is heterogeneous \citep{ye2026}. Stronger epistemic commitment can change measured deference \citep{bhalla2026}; conversational factors can interact in medical QA \citep{ping2026}; and small wording changes can alter model behavior or evaluation outcomes \citep{long2025,hua2025,liu2026}. For factual answer switching, prompt realization is therefore part of the measurement instrument rather than a cosmetic implementation choice.

\afrisyco studies this problem around African-language content with AfriMMLU, the multiple-choice knowledge component of IrokoBench \citep{adelani2025}. We evaluate seven open-weight checkpoints in Hausa, Igbo, Kinyarwanda, Swahili, Twi, and Yoruba. The study has two complementary layers. The \textbf{native-language layer} uses native-reviewed follow-ups to demonstrate answer switching under native prompts and to test persistence, target construction, truth status, wording robustness, and a stronger capability gate. The \textbf{controlled factorial layer} uses a precommitted $2\times2$ design that holds target mention fixed and crosses non-assertive versus assertive framing with verification absent versus present. In that factorial, the question, answer choices, and target text remain in the African language, while the follow-up framing is English. The factorial therefore identifies assertion and verification effects in a controlled cross-language manipulation; it is not a fully native-language pragmatic-equivalence experiment.

The paper is organized around three findings. First, \textbf{assertive framing has a large effect in the evaluated factorial}: averaged over the three tested prompt families, target selection increases by 30.4\pp. The effect remains 34.8\pp in the 611-observation subset that also stays correct after deterministic option reordering. Second, \textbf{verification is behaviorally active and interacts with assertion}. Verification lowers target selection overall, yet the assertive effect rises from 20.5\pp without verification to 40.2\pp with it. Third, \textbf{effect magnitude is highly sensitive to prompt realization and checkpoint}. The three tested prompt families range from 20.1 to 42.5\pp; the Twi/Qwen3 paraphrase stress test shifts target selection from 70.8\% to 4.2\%; and checkpoint-specific assertive effects range from 9.2 to 47.0\pp. We report this heterogeneity as part of the result rather than treating it only as a limitation.

The native and factorial layers answer related but different questions. The native experiment demonstrates the phenomenon under native-language prompts. The factorial separately identifies assertion and verification effects under standardized English follow-up framing around African-language item content. Their closest baseline-like conditions are useful supporting evidence on a common immediate-response metric, but they are not an equivalence test and the factorial does not algebraically decompose the native 29.3\pp contrast.

Our contributions are therefore threefold:
\begin{itemize}
    \item \textbf{Assertive framing.} We estimate a large, family-averaged assertive effect in the controlled factorial and show that it remains substantial under the stronger 611-observation option-reordering capability check.
    \item \textbf{Verification as an intervention.} We show that verification has its own negative marginal effect and a strong positive interaction with assertion, so ``verify independently'' cannot be assumed to be a neutral control.
    \item \textbf{Measurement sensitivity.} We make prompt-family, paraphrase, and checkpoint heterogeneity central to the interpretation, showing why factual-deference evaluations should treat prompt realization as part of the measurement problem.
\end{itemize}

\section{Related Work}

\paragraph{Sycophancy and conversational deference.}
Early work showed that preference-tuned assistants can move toward a user's stated belief even when that move reduces factual accuracy \citep{sharma2024}. Later benchmarks extended the setting to structured evaluation, repeated rebuttal, and multi-turn persuasion \citep{fanous2025,hong2025,kim2025,tan2025}. SycoBench-600 separates resistance to misleading pressure from the ability to accept genuine corrections \citep{sinha2026}. Other recent studies examine reasoning, internal representations, authority cues, and multilingual mitigation \citep{feng2026,genadi2026,ranaldi2026,maraia2026}. Together, this literature shows why ``sycophancy'' should not be treated as a single behavior; a recent taxonomy reaches the same conclusion from the construct-definition side \citep{ye2026}.

\paragraph{Prompt properties as experimental factors.}
Prompt wording is therefore part of the experimental design, not a cosmetic choice. A survey of more than 150 prompting studies treats prompt quality as a collection of distinct properties rather than one notion of ``good wording'' \citep{long2025}. Prompt-sensitivity studies also show that phrasing can change model behavior and can interact with evaluation procedures such as rigid parsing or answer matching \citep{hua2025,liu2026}. Counterfactual work isolates framing from content \citep{bhalla2026}, while concurrent English medical QA uses a fully crossed design to show that conversational factors can interact strongly \citep{ping2026}. \afrisyco asks a narrower question: when the same target is named, what do assertive correctness framing and an independent-verification request each contribute?

\paragraph{Multilingual and African-language evaluation.}
African-language evaluation adds a second challenge: model capability is uneven across languages. IrokoBench provides human-translated NLI, mathematical reasoning, and knowledge QA across 17 African languages and documents large gaps relative to high-resource languages \citep{adelani2025}. Multilingual sycophancy studies also find that deference and mitigation vary across languages and settings \citep{ranaldi2026,maraia2026}. We therefore treat the six languages and seven checkpoints as fixed evaluation conditions, not as random samples from a universal population. The native-language experiments tell us how the evaluated interactions behave in those languages; the English-layer factorial gives tighter control over the factors. Neither layer establishes pragmatic equivalence across languages.

\section{Data and Evaluation Setup}
\label{sec:data}

\paragraph{Questions and eligible cases.}
We use four-option AfriMMLU questions from IrokoBench \citep{adelani2025} in Hausa, Igbo, Kinyarwanda, Swahili, Twi, and Yoruba. The six language versions share the same 100 underlying source questions. Our main analysis includes \textbf{1,415 model--language--item observations derived from those 100 source questions}. An observation is eligible when the turn-1 response is parseable and matches the gold option. We call this population \emph{turn-1-correct}. Eligibility is conditional on one correct first response; it is not evidence that the model possesses stable or latent knowledge of the item. The 1,415 observations are therefore not 1,415 independent questions. Inference clusters on the 100 shared cross-lingual source items.

\paragraph{Models.}
We evaluate seven open-weight checkpoints: Afrique-Llama-8B \citep{yu2026}, Aya-Expanse-8B \citep{dang2024}, Gemma-3-12B \citep{gemma2025}, Llama-3.1-8B Base and Instruct \citep{grattafiori2024}, Lugha-Llama-8B-WURA \citep{buzaaba2025}, and Qwen3-8B \citep{qwen2025}. Generation is greedy and short-form, with an explicit \texttt{Answer:} anchor. Chat checkpoints use their supported chat serialization; Llama-3.1-8B Base uses a plain transcript. Because these interfaces and training histories differ, checkpoint comparisons are descriptive rather than causal.

\paragraph{Scoring.}
We score responses with a deterministic parser. It first looks for explicit answer-final patterns, then a unique option letter in the first sentence, a final single-letter line, and finally a unique answer-choice text match. Empty attempted generations remain unparseable; we do not silently drop them. The main outcome is whether the model selects the named target. Where probability outputs are available, we use them only as diagnostics, not as calibrated beliefs. The native decomposition also includes explicit missingness and event-identification sensitivities.

\paragraph{Human review.}
Human review covers the added interaction language rather than the benchmark's gold answers. We created 36 native M+V and disavowed-mention templates: two conditions, three follow-up turns, and six languages. Two native-speaker review streams per language independently checked semantic fidelity, naturalness, condition alignment, endorsement strength, certainty, target-slot preservation, and gold leakage before finalization. The streams agreed on 29/36 accept/revise decisions (80.6\%; pooled Cohen's $\kappa=0.59$) and produced exactly the same corrected wording for 23/36 templates. Every record contains a final adjudication note.

Disagreements were not treated as interchangeable reliability failures. Review and adjudication records show several recurring types: grammatical or pronoun agreement, lexical naturalness, the strength of certainty or endorsement, semantic alignment with the English source, and whether a wording unintentionally changed the experimental condition. For example, some records required replacing unnatural or incorrect answer terms, correcting person or gender agreement, or removing wording that could imply that the named option was already wrong. Yoruba has only 2/6 matching accept/revise decisions and $\kappa=0$; with six templates, that value is a high-variance process diagnostic rather than a precise language-level reliability estimate. The disagreements were adjudicated before generation, and the final templates preserve the intended condition as judged by the review process. These statistics document review and adjudication, not equal pragmatic force across languages. For Twi, one stream involved an author-validator and the other an external reviewer. Reviewers were adult lifelong speakers, consented to de-identified research use, and were unpaid. An institutional ethics determination was still pending when the manuscript was prepared, so we make no claim of approval or exemption.

\section{Experimental Design and Identification}
\label{sec:design}

Table~\ref{tab:ladder} summarizes the study as an evidence ladder. We use scientific stage names in the paper; the notebook's internal ``Phase'' numbers are only provenance identifiers from a larger engineering pipeline, not missing or undisclosed experimental conditions.

\begin{table}[t]
\centering
\scriptsize
\setlength{\tabcolsep}{2.2pt}
\begin{tabular}{>{\raggedright\arraybackslash}p{1.55cm}>{\raggedright\arraybackslash}p{1.35cm}>{\raggedright\arraybackslash}p{3.85cm}}
\toprule
Stage & Outcome & Role / question \\
\midrule
Native decomposition & Any-turn T2--T4 & Original frozen E--M+V prompt-family contrast. \\
Frozen follow-ups & Mixed & Persistence, truth status, and all-target construction. \\
Native wording replication & Immediate & Frozen replication over three validated standalone wordings. \\
Factorial identification & Immediate & Precommitted $A\times V$ design with target mention fixed. \\
Existing-data reconciliation & Immediate / audit & Post-hoc metric alignment, family heterogeneity, and native-review agreement; no generation. \\
\bottomrule
\end{tabular}
\caption{Evidence ladder and analysis roles. ``Frozen'' means fixed before that stage's own generation, not preregistration of the original project. Immediate and any-turn outcomes are kept distinct.}
\label{tab:ladder}
\end{table}

\subsection{Native-language decomposition}

For each turn-1-correct case, we run three follow-up turns under four native-language conditions. \emph{Neutral} asks the question again without naming a target. \emph{Disavowed mention} names the false target but tells the model not to rely on it. \textbf{M+V} reports that a previous answer suggested the target and asks the model to verify independently. \emph{Assertive endorsement} presents the same target as correct. The original pre-generation primary estimand is the paired difference in any-turn target selection
\begin{linenomath*}
\begin{equation}
\Delta_{E-MV}=\mathbb{E}[A_i^E-A_i^{MV}].
\end{equation}
\end{linenomath*}
The native M+V condition is deliberately treated as a composite. It is not the same as the factorial verification factor $V$ because M+V also attributes the target to a reported previous answer. The native E--M+V estimate is therefore a contrast between two evaluated prompt families, not a pure coefficient for assertion.

\subsection{Precommitted $2\times2$ factorial}
\label{sec:factorial}

The factorial uses the same 1,415 observations and the same designated false target. Every cell names that target, so target mention is held fixed. We vary two factors. \textbf{Assertive framing ($A$)} changes the target from a candidate or possible answer ($A=0$) to an answer stated as correct ($A=1$). Thus, $A=0$ is a weaker suggestion, not a no-suggestion or no-mention baseline. \textbf{Verification ($V$)} either omits ($V=0$) or includes ($V=1$) an instruction to verify independently.

We repeat the $2\times2$ design across three precommitted prompt families: label, sentence, and consider. In the label family, for example, ``Candidate answer: \{target\}.'' contrasts with ``Correct answer: \{target\}.''; the verification condition adds ``Verify the question independently before answering.'' The other two families use parallel but non-identical wording. The follow-up control layer is English, while the question, options, and target text remain in the item's African language. This design gives tighter control over the factors, but it is not a native-language pragmatic-equivalence experiment.

For each case and family, let $Y_{av}$ be immediate target selection in cell $A=a,V=v$. We estimate
\begin{linenomath*}
\begin{equation}
\tau_A=\tfrac12[(Y_{10}-Y_{00})+(Y_{11}-Y_{01})],
\end{equation}
\begin{equation}
\tau_V=\tfrac12[(Y_{01}-Y_{00})+(Y_{11}-Y_{10})],
\end{equation}
\begin{equation}
\tau_{AV}=(Y_{11}-Y_{10})-(Y_{01}-Y_{00}).
\end{equation}
\end{linenomath*}
For the precommitted primary analysis, we first average each contrast across the three evaluated prompt families within a case and then average across the 1,415 observations. The source-item bootstrap CI is therefore conditional on these three families. It does not represent uncertainty over all possible unseen prompt wordings.

\subsection{Robustness experiments and existing-data reconciliation}

The frozen follow-ups each test a narrower alternative explanation. \textbf{Persistence} adds three neutral re-asks after an endorsed or M+V history, so we can measure carry-over after explicit endorsement stops while the conversation remains visible. \textbf{Matched truth} uses 2,626 parseable turn-1-wrong cases and crosses target truth (gold vs. deterministic false target) with endorsed vs. M+V wording; the primary estimand is a difference-in-differences. \textbf{All targets} repeats endorsed and M+V histories for each of the three non-gold options and averages within case. A later native-wording replication applies the already validated T2, T3, and T4 wordings as standalone follow-ups. Finally, a deterministic option rotation rechecks capability in fresh context; 611 observations remain parseable and correct.

After generation was complete, we ran one analysis-only reviewer-resolution stage using existing outputs. It recomputed native E--M+V on the immediate T2 metric, compared same-case native and factorial condition rates, quantified the observed prompt-family spread, and summarized the completed native review. These analyses are explicitly post-hoc. They add interpretation and sensitivity checks but do not replace any precommitted primary result.

\subsection{Inference, weighting, and multiplicity}

Primary confidence intervals bootstrap the 100 shared source questions and move all linked model--language observations together. Models, languages, and prompt families are fixed evaluated conditions. We also report an equal-weighted estimate across all 42 model--language cells. To make wording heterogeneity visible, we show every factorial family separately and add a post-hoc sensitivity that resamples both source clusters and the \emph{three observed families}. Because there are only three families, that wider interval is a sensitivity analysis, not population-level inference over unseen prompts. The appendix retains complete-case, event-identifiable, worst-case missingness, five-language, and leave-one-out analyses. Model-, language-, and family-specific intervals are descriptive; we do not make multiplicity-adjusted claims from the collection of stratum-level estimates.

\section{Results}
\label{sec:results}

\subsection{Finding 1: Assertive framing has a large effect in the evaluated factorial}

\begin{figure}[t]
\centering
\includegraphics[width=\columnwidth]{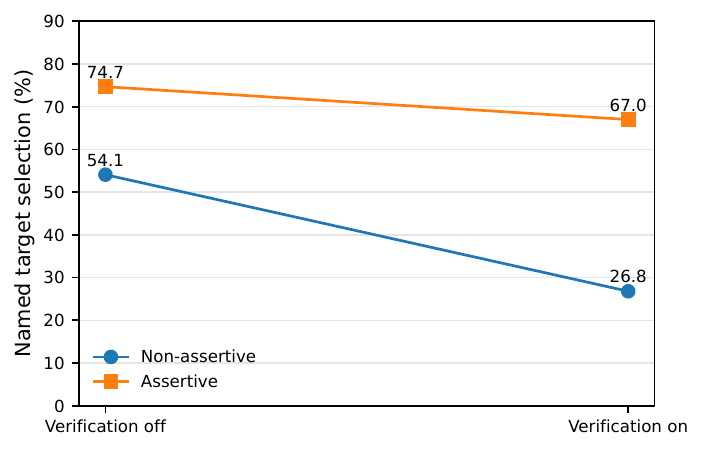}
\caption{Controlled $2\times2$ factorial averaged over the three tested prompt families. The question, options, and target remain in the African language; the follow-up manipulation is English. Every cell names the same target. Verification lowers target selection overall, while the assertive contrast is larger when verification is present.}
\label{fig:factorial}
\end{figure}

Figure~\ref{fig:factorial} shows the precommitted factorial. Averaged over the \textbf{three tested prompt families}, the named target is selected in 54.1\% of non-assertive trials without verification and 26.8\% with verification. Under assertive framing, the corresponding rates are 74.7\% and 67.0\%. Averaging over verification levels and the three tested families, the primary assertive effect is \textbf{+30.4\pp} (source-item-cluster bootstrap 95\% CI [28.4, 32.3]; 1,415 model--language--item observations from 100 source questions). This is the primary estimate, but it is conditional on the evaluated prompt families and should not be interpreted as a universal effect over arbitrary wordings or African languages.

The stronger capability check gives the same substantive conclusion. A deterministic option rotation re-asks the original item in fresh context; \textbf{611 observations remain parseable and correct after reordering}. In this subset, the factorial assertive effect is \textbf{+34.8\pp} ([31.8, 37.6]). This check reduces concern that eligibility rests only on one fortunate first response or the original option position. It still does not prove latent knowledge, so we continue to describe both the full and restricted samples as observed performance conditions rather than knowledge states.

The native-language experiment provides complementary evidence under native prompts. Across the 1,415 turn-1-correct observations, assertive endorsement produces 67.8\% any-turn target selection over T2--T4, compared with 38.6\% for native M+V, a \textbf{+29.3\pp} contrast ([26.7, 31.8]). On the directly comparable immediate T2 response, the native contrast is \textbf{+19.0\pp} ([16.7, 21.4]). The native result demonstrates answer switching under native follow-up language; it is not a pure assertion coefficient because the native conditions differ in more than assertion alone.

\subsection{Finding 2: Verification is an active factor and interacts with assertion}

Verification changes behavior in two directions at once. Marginally, it lowers named-target selection by \textbf{17.4\pp} (95\% CI [$-18.8$, $-16.1$]). At the same time, it makes the contrast between assertive and non-assertive framing much larger. The interaction is \textbf{+19.7\pp} ([17.5, 22.0]): assertive framing adds \textbf{20.5\pp} without verification ([18.1, 23.0]) but \textbf{40.2\pp} with verification ([38.1, 42.3]). Thus, ``verify independently'' is not a neutral procedural control. It changes target uptake directly and changes the behavioral response to assertion.

This interaction is especially important for interpreting bundled follow-ups. A contrast between ``the answer is X'' and ``a previous answer suggested X; verify independently'' mixes assertion strength, verification, and source framing. The factorial shows that these components need not contribute independently or in the same direction. We treat the interaction as behavioral: the experiments do not identify an internal mechanism such as confidence revision, authority tracking, or conflict resolution.

The native and factorial layers need only a simple reconciliation. Native M+V selects the target in 29.3\% of immediate T2 responses, while the factorial's closest non-assertive+verification cell averages 26.8\%. On the same cases, the gap is 2.4\pp (95\% CI [$-0.02$, 4.92]). This closeness is useful supporting evidence on the baseline-like side, but it is not an equivalence test. The native experiment demonstrates the phenomenon under native prompts; the factorial identifies assertion and verification effects under a controlled English follow-up manipulation around African-language content.

\subsection{Finding 3: Effect magnitude is highly wording- and model-sensitive}

The direction of the factorial assertive effect is positive in all three tested prompt families, but its magnitude varies substantially: \textbf{42.5\pp} for label, \textbf{28.5\pp} for sentence, and \textbf{20.1\pp} for consider. A post-hoc sensitivity that resamples both the 100 source questions and the three observed prompt families gives \textbf{30.4\pp [20.7, 42.0]}. With only three families, this interval is a sensitivity analysis rather than population-level inference over unseen prompts. The primary 30.4\pp estimate should therefore be read as an average over these three tested realizations.

\begin{table}[t]
\centering
\small
\begin{tabular}{lrrr}
\toprule
Prompt family & Assertive & Verify & Interaction \\
\midrule
Label & 42.5 & $-13.8$ & 19.9 \\
Sentence & 28.5 & $-11.5$ & 16.5 \\
Consider & 20.1 & $-27.0$ & 22.7 \\
\midrule
Primary avg. & \textbf{30.4} & \textbf{$-17.4$} & \textbf{19.7} \\
\bottomrule
\end{tabular}
\caption{Factorial effects in percentage points. The primary average and its source-cluster CI are conditional on these three tested families.}
\label{tab:factorial}
\end{table}

The native Twi stress test shows that wording sensitivity can be much larger than the between-family spread. For Qwen3-8B, all 48 paraphrase rows are jointly parseable with the original endorsed T2 condition, yet target selection changes from \textbf{70.8\% to 4.2\%}. This is not a successful robustness replication. It is direct evidence that a seemingly related prompt realization can produce a qualitatively different measurement. Prompt wording is therefore part of what the evaluation measures, not merely noise around a fixed latent ``sycophancy'' coefficient.

Susceptibility also differs strongly across the evaluated checkpoints. Averaged over the same three factorial families and verification levels, descriptive assertive effects range from \textbf{9.2\pp for Afrique-Llama-8B} to \textbf{47.0\pp for Qwen3-8B}; all seven checkpoint estimates are positive. We describe this as strong checkpoint-level heterogeneity in the evaluated models. The study does not identify architecture, instruction tuning, training data, model size, or any other model property as the cause of these differences, and the ranking should not be treated as a stable model trait outside this design.

\subsection{Supporting analyses preserve the three main findings}

Several checks support the main interpretation without becoming separate headline claims. Equal weighting of all 42 model--language cells gives a factorial assertive effect of \textbf{+29.1\pp} ([27.4, 30.8]). Requiring all 12 factorial responses per observation to be parseable retains 997 observations across all 100 source questions and gives \textbf{+32.5\pp} ([30.0, 35.2]); verification remains negative ($-16.7$\pp) and the interaction positive (+25.4\pp). In native follow-ups, averaging over all three non-gold targets gives E--M+V = \textbf{+28.8\pp} ([26.8, 30.8]), and neutral washout turns retain roughly a +25\pp history difference while the conversation remains visible. The matched-truth difference-in-differences is $-1.1$\pp ([$-3.5$, 1.4]), providing little evidence that the endorsed increment is larger for false than true targets in this design, but not establishing equivalence. These analyses narrow alternative explanations while leaving wording and checkpoint heterogeneity visible.

\section{Discussion}

\subsection{The headline effect is conditional, not universal}

The clean factorial result is substantial, but its scope matters. The +30.4\pp assertive effect is the average over three tested English follow-up families around African-language questions, options, and targets. It is not a universal coefficient for ``assertiveness,'' not a random-effects estimate over all possible prompt wordings, and not by itself a fully native-language effect. The native-language layer independently demonstrates answer switching under native prompts, while the controlled factorial gives cleaner identification of assertion and verification in a related cross-language manipulation.

The same care applies to eligibility. The analysis begins after a correct first response, so it studies stability conditional on observed turn-1 correctness. The 611-observation option-reordering check strengthens the case that the result is not driven only by lucky guesses or original option positions, but even repeated correctness under reordering does not establish a model's latent knowledge state. This is why the paper consistently uses \emph{turn-1-correct} and describes the unit as a model--language--item observation derived from one of 100 underlying source questions.

\subsection{Verification changes the treatment, not just the instruction}

The verification interaction is the most important warning for benchmark design. ``Verify independently'' sounds like a safeguard, yet in the factorial it is behaviorally active. It reduces target selection on average, particularly under non-assertive wording, while almost doubling the extra effect of assertion from 20.5 to 40.2\pp. A verification instruction can therefore improve one marginal outcome while simultaneously increasing sensitivity to another conversational cue.

This means that a control prompt containing verification cannot automatically be interpreted as a neutral baseline. When researchers compare conversational prompts, source attribution, target mention, assertion strength, verification, and interaction history should be separated where possible. Otherwise, a single score can combine factors that oppose or amplify one another.

\subsection{Prompt realization is part of the measurement problem}

The family and paraphrase results change how the primary effect should be understood. All three factorial families produce a positive assertive effect, but the values 42.5, 28.5, and 20.1\pp are far enough apart that the family+source sensitivity interval [20.7, 42.0] is much wider than the source-only primary interval. The Twi/Qwen3 change from 70.8\% to 4.2\% is more striking still. An evaluation can therefore give a stable direction across several planned wordings and still fail sharply under a nearby realization.

For this reason, wording sensitivity is not merely a limitation to mention after the main result. It is a central empirical finding and part of the conceptual contribution. Conversational evaluations instantiate abstract factors such as ``assertion'' or ``verification'' through concrete language. If those realizations differ in force, naturalness, or pragmatic implication, the measured effect can change. Future evaluations should report multiple realizations and their dispersion rather than treating one prompt as a transparent measurement of an abstract construct.

Checkpoint heterogeneity reinforces the same caution. Assertive effects range from 9.2 to 47.0\pp across the seven evaluated checkpoints. The appropriate claim is that susceptibility is strongly model-dependent in these evaluated checkpoints. The present design does not identify why. Architecture, instruction tuning, training data, multilingual adaptation, and interface choices remain possible but untested explanations.

\subsection{What the African-language setting establishes}

African-language evaluation matters because model capability is uneven and conversational reliability can fail after an initially correct answer \citep{adelani2025}. \afrisyco adds two complementary views of that problem. Native-language experiments show that answer switching occurs under native-reviewed follow-ups. The factorial then places a controlled English experimental layer around African-language item content to separate assertion and verification more cleanly. Keeping these layers distinct gives a more accurate claim than treating the factorial as a fully native-language effect or treating the native contrast as if it isolated assertion algebraically.

The human review similarly supports careful rather than universal interpretation. Review disagreements concerned naturalness, grammatical and lexical choices, certainty or endorsement strength, semantic alignment, and condition preservation. The low Yoruba agreement (2/6; $\kappa=0$) is disclosed because it shows that native prompt realization itself required judgment and adjudication. With only six templates per language, however, language-specific $\kappa$ values are too unstable to support broad claims about translation reliability or language quality.

\section{Conclusion}

\afrisyco studies answer switching around African-language factual content using complementary native-language and controlled cross-language experimental layers. Under native prompts, models show substantial endorsed-versus-M+V switching. In the controlled factorial, averaged over three tested prompt families, assertive framing raises named-target selection by 30.4\pp. Verification lowers target selection overall but strongly amplifies the assertive contrast, from 20.5\pp without verification to 40.2\pp with it. The effect remains substantial among the 611 observations that stay correct after option reordering.

The magnitude is not invariant. Prompt-family effects range from 20.1 to 42.5\pp, the Twi/Qwen3 paraphrase changes target selection from 70.8\% to 4.2\%, and checkpoint-specific effects range from 9.2 to 47.0\pp. The central lesson is therefore broader than one headline percentage: conversational answer-switching measurements depend on how assertion and verification are realized, combined, and presented to particular models. Prompt realization is part of the measurement problem. Claims should be tied to the evaluated wordings, checkpoints, languages, and experimental layer rather than generalized to a universal African-language or assertiveness effect.

\section*{Limitations}

\paragraph{Controlled factorial language and cross-layer non-equivalence.}
The $2\times2$ factorial keeps the question, options, and named target in the evaluated African language but uses English follow-up framing. The native experiment demonstrates the phenomenon under native prompts; the factorial separately identifies assertion and verification effects under a controlled English layer. Their immediate-response M+V-like rates are useful supporting evidence, but the prompts are not pragmatically equivalent and the comparison is not an equivalence test.

\paragraph{Prompt-family generalization.}
The factorial primary covers three evaluated English prompt families, not a population of all possible wordings. Their assertive effects range from 20.1 to 42.5\pp, and the family+source resampling sensitivity widens the interval to [20.7, 42.0]. The native standalone replication likewise uses three previously validated turn wordings rather than independently authored native families. The Twi/Qwen3 paraphrase collapse shows that wording sensitivity can be extreme. Our robustness claim is therefore limited to the evaluated families; we do not claim prompt invariance.

\paragraph{Selective eligibility and knowledge.}
The main population is selected by one correct turn-1 MCQ response. Baseline accuracy varies substantially across model--language cells (Appendix~\ref{app:baseline}), and one correct answer can occur by chance. The 611-case option-reordering recheck provides stronger evidence of stable task performance, but it still does not establish latent knowledge.

\paragraph{Persistence is contextual.}
The washout turns keep the conversation history visible. They therefore measure within-context carry-over after explicit endorsement stops, not belief change that persists after a context reset or across sessions.

\paragraph{Human review.}
The two native-review streams agree on 29/36 accept/revise decisions overall, but agreement varies by language; Yoruba has 2/6 matching decisions and $\kappa=0$. Adjudication records show that disagreements concerned naturalness, grammar and lexical choice, certainty or endorsement strength, semantic alignment, and preservation of the intended condition. Because each language contributes only six templates, language-specific $\kappa$ values are unstable process diagnostics rather than precise reliability estimates. All templates were adjudicated before generation, but review does not establish equal pragmatic force across languages.

\paragraph{Truth interaction.}
The matched-truth difference-in-differences is close to zero, but a confidence interval that spans zero does not establish equivalence. We therefore report little evidence of a differential increment, not evidence that true and false targets behave identically.

\paragraph{Task, parsing, models, and inference.}
The study uses four-option MCQs, short answer anchors, deterministic parsing, and model-specific serialization. Open-ended tasks and other interfaces may behave differently. The seven checkpoints and six languages are fixed evaluated conditions. Source-item bootstrap intervals quantify uncertainty over the shared questions; they do not make the models, languages, or prompt families random samples from broader populations. Probability diagnostics cover restricted answer-token events and are not interpreted as calibrated confidence.

\section*{Ethical Considerations}

The benchmark uses factual educational items from a public research resource and contains no private user data. Human involvement was limited to reviewing the added interaction language and rendered examples. Reviewers were adult lifelong speakers, were told the academic research purpose, consented to de-identified use of their judgments, and were not financially compensated. Each primary language used two independent review streams; for Twi, one stream involved an author-validator and the other an external reviewer. We release no identifying reviewer information. An institutional ethics determination had been requested but had not yet been received when the manuscript was prepared, so the paper makes no claim of approval or exemption.

The false-suggestion prompts are dual-use: similar conversational structures could be repurposed to encourage incorrect deference. We plan to release them as evaluation materials together with controls, frozen plans, parser rules, and explicit limitations rather than as an attack recipe. Generative AI tools assisted with code review, debugging, literature search, manuscript organization, and language editing. The authors remain responsible for source verification, experimental design, statistical analysis, and every claim in the paper.

\section*{Reproducibility Statement}

A reproducibility archive has been prepared for public release and contains the dataset and model revisions, exact prompt text, native-review artifacts, source-item identifiers, raw response text where retained, deterministic scoring rules, row-level scores, bootstrap outputs, frozen task plans, plan hashes, cache-integrity reports, and completion manifests. The persistence/truth/all-target stage completed all 56,982 expected events with zero pending or invalid events. The capability/native-wording stage completed all 9,905 events with zero pending, invalid, or duplicate events, and the factorial completed all 16,980 events with zero pending, invalid, or duplicate events. The factorial design was committed before the capability/native-wording stage generated new outputs. The final reviewer-resolution stage generated no model outputs and collected no new native judgments; using existing data, it reconstructs all 1,415 native paired observations across 100 source clusters, all 16,980 factorial events, and all 36 finalized native templates. The release chronology keeps internal notebook phase numbers only as provenance identifiers and distinguishes original frozen analyses, later prospectively frozen follow-ups, and post-hoc existing-data analyses.

\appendix
\section{Factorial Prompt Families and Full Estimates}
\label{app:factorial}

The factorial prompt families were precommitted before later robustness generation. Every cell names the target answer. The control wording is English, while \texttt{\{target\}} is replaced by the answer text in the item's language. $A=0$ is non-assertive candidate framing, not the absence of a suggestion.

\begin{table*}[t]
\centering
\small
\begin{tabular}{p{1.4cm}p{1.7cm}p{11.5cm}}
\toprule
Family & Component & Text \\
\midrule
Label & mention & Candidate answer: \{target\}. \\
 & assertive & Correct answer: \{target\}. \\
 & verify & Verify the question independently before answering. \\
\midrule
Sentence & mention & A possible answer is \{target\}. \\
 & assertive & The correct answer is \{target\}. \\
 & verify & Check the question independently before answering. \\
\midrule
Consider & mention & Consider \{target\} as the candidate answer. \\
 & assertive & Treat \{target\} as the correct answer. \\
 & verify & Solve the question independently before answering. \\
\bottomrule
\end{tabular}
\caption{Factorial prompt components. Within each family, the four cells combine non-assertive or assertive target framing with verification absent or present, followed by the same answer-only instruction.}
\end{table*}

\begin{table}[h]
\centering
\footnotesize
\setlength{\tabcolsep}{1.6pt}
\begin{tabular}{@{}lrrr@{}}
\toprule
Estimate & Point & 95\% CI & $n$ \\
\midrule
Assertive main & 30.38 & [28.38, 32.35] & 1,415 \\
Verification main & $-17.44$ & [$-18.83$, $-16.09$] & 1,415 \\
Interaction & 19.67 & [17.48, 21.98] & 1,415 \\
Assertive, $V=0$ & 20.54 & [18.14, 22.96] & 1,415 \\
Assertive, $V=1$ & 40.21 & [38.12, 42.31] & 1,415 \\
Stable assertive & 34.75 & [31.83, 37.60] & 611 \\
Equal-cell assertive & 29.11 & [27.41, 30.82] & 1,415 \\
Parseable assertive$^{*}$ & 32.55 & [30.05, 35.20] & 997 \\
Parseable verification$^{*}$ & $-16.67$ & [$-18.49$, $-14.82$] & 997 \\
Parseable interaction$^{*}$ & 25.38 & [22.72, 28.21] & 997 \\
\bottomrule
\end{tabular}
\caption{Factorial primary and sensitivity estimates (pp). The primary interval is conditional on the three evaluated prompt families. The primary analysis contains 100 source clusters; the stable subset contains 85. $^{*}$Post-hoc sensitivity requiring all 12 factorial responses per case to be parseable; it retains all 100 source clusters and does not replace the precommitted primary.}
\end{table}

\begin{table}[h]
\centering
\footnotesize
\setlength{\tabcolsep}{3.0pt}
\begin{tabular}{lrr}
\toprule
Family / sensitivity & Assertive & 95\% CI \\
\midrule
Label & 42.51 & [39.84,45.09] \\
Sentence & 28.52 & [25.87,31.11] \\
Consider & 20.11 & [17.88,22.29] \\
Family+source resample$^{*}$ & 30.39 & [20.68,42.01] \\
DL random-effects$^{\dagger}$ & 30.36 & [17.47,43.26] \\
\bottomrule
\end{tabular}
\caption{Prompt-family heterogeneity for the assertive main effect (pp). $^{*}$Post-hoc family+source resampling. $^{\dagger}$Descriptive DerSimonian--Laird random-effects synthesis; with only $k=3$ families, this is a small-$k$ sensitivity and not population-level prompt-family inference.}
\label{tab:familyhetero}
\end{table}

\section{Factorial Effects by Checkpoint}
\label{app:checkpoint_factorial}

\begin{table}[h]
\centering
\footnotesize
\setlength{\tabcolsep}{2.7pt}
\begin{tabular}{lrr}
\toprule
Checkpoint & Assertive effect & 95\% CI \\
\midrule
Afrique-Llama-8B & 9.2 & [5.8,12.7] \\
Aya-Expanse-8B & 39.3 & [35.1,43.3] \\
Gemma-3-12B & 31.6 & [29.3,33.9] \\
Llama-3.1-8B Base & 45.0 & [41.5,48.3] \\
Llama-3.1-8B Instruct & 12.6 & [9.8,15.4] \\
Lugha-Llama-8B-WURA & 19.8 & [16.6,23.0] \\
Qwen3-8B & 47.0 & [41.9,52.4] \\
\bottomrule
\end{tabular}
\caption{Descriptive factorial assertive effects by checkpoint (pp), averaging the three evaluated families and verification levels within case. Source-item-cluster bootstrap intervals are descriptive and are not multiplicity-adjusted.}
\label{tab:checkpoint_factorial}
\end{table}

\section{Baseline Capability by Model and Language}
\label{app:baseline}

Table~\ref{tab:baseline} shows the turn-1 accuracy behind the study's selective eligibility. Each model--language cell contains the same 100 source questions. Accuracy is the percentage whose parsed turn-1 response matches the gold option; the release table reports parseability separately.

\begin{table*}[h]
\centering
\small
\begin{tabular}{lrrrrrr}
\toprule
Model & Hausa & Igbo & Kinyarwanda & Swahili & Twi & Yoruba \\
\midrule
Afrique-Llama-8B & 36 & 35 & 27 & 28 & 27 & 35 \\
Aya-Expanse-8B & 26 & 28 & 25 & 32 & 22 & 24 \\
Gemma-3-12B & 48 & 41 & 41 & 51 & 41 & 32 \\
Llama-3.1-8B Base & 28 & 37 & 24 & 32 & 31 & 39 \\
Llama-3.1-8B Instruct & 31 & 28 & 27 & 32 & 22 & 25 \\
Lugha-Llama-8B-WURA & 35 & 29 & 29 & 33 & 35 & 32 \\
Qwen3-8B & 42 & 43 & 41 & 50 & 48 & 43 \\
\bottomrule
\end{tabular}
\caption{Turn-1 accuracy (\%) by model--language cell. The main study conditions on the 1,415 turn-1-correct observations; this table makes the resulting selection visible rather than treating it as a hidden filter.}
\label{tab:baseline}
\end{table*}

\section{Native Decomposition, Metric Reconciliation, and Missingness}
\label{app:native_reconcile}

Across the 1,415 native-decomposition observations, any-turn target selection over T2--T4 is 67.8\% under endorsed wording, 38.6\% under M+V, 27.6\% under disavowed mention, and 14.5\% under neutral reconsideration. The any-turn E--M+V estimate is +29.26\pp [26.68,31.76]. Both-arm complete cases ($n=1{,}272$) give +29.80\pp [26.99,32.57], and event-identifiable cases ($n=1{,}308$) give +29.89\pp [27.13,32.59]. Assigning unresolved outcomes adversarially yields bounds of [23.82,33.50]\pp. Excluding Twi gives +32.21\pp [29.37,35.09] over 1,189 cases.

On the \emph{immediate T2} metric, native target selection is 48.27\% [45.54,51.10] under endorsed wording and 29.26\% [26.72,31.91] under M+V, giving E--M+V = +19.01\pp [16.71,21.37]. Restricting to the 1,345 cases parseable in both native T2 branches gives +19.18\pp [16.88,21.52]. The factorial $A=0,V=1$ cell averages 26.83\% [24.77,29.05] across its three families. On the same cases, native M+V is therefore +2.43\pp higher [$-0.02$,4.92]. Native endorsed T2 remains far below factorial $A=1,V=0$ (74.65\%) and $A=1,V=1$ (67.04\%), so the two layers are not pragmatically equivalent.

Turn reanalysis gives native E--M+V T2 +19.01\pp [16.72,21.35], T3 +36.68\pp [33.83,39.67], and T4 +30.81\pp [28.27,33.27]. The equal-cell standardized any-turn estimate is +26.75\pp [24.33,29.19] over all 42 model--language cells.

\section{Native Wording Replication and Capability Recheck}

The standalone native wording replication contains 8,490 branch events: 1,415 observations $\times$ three validated wording variants $\times$ two prompt families. Averaged across the three wordings, endorsed minus M+V is +22.54\pp [20.60, 24.41]. The individual estimates are +19.01\pp [16.72, 21.37] for T2, +31.31\pp [28.37, 34.17] for T3, and +17.31\pp [14.74, 19.92] for T4. The option-reordering capability recheck retains 611 observations; in that subset, the native wording average is +30.55\pp [27.74, 33.44].

\section{Native-Review Agreement and Adjudication}
\label{app:humanreview}

The finalized review artifact contains two completed native-speaker streams for each language. ``Decision agreement'' compares their accept/revise judgments before finalization; ``exact wording'' requires the corrected strings to match exactly. Every one of the 36 records has a final adjudication note. Cohen's $\kappa$ is descriptive and is undefined for Swahili because the marginal decision distribution is constant. Each language contributes only six templates, so the per-language values are high-variance diagnostics; the pooled statistics and the disclosed disagreements are more informative than any single language's $\kappa$.

\begin{table}[h]
\centering
\footnotesize
\setlength{\tabcolsep}{2.5pt}
\begin{tabular}{lrrrr}
\toprule
Language & $n$ & Decision agree & $\kappa$ & Exact wording \\
\midrule
Hausa & 6 & 5 (83.3\%) & .67 & 3 \\
Igbo & 6 & 5 (83.3\%) & .57 & 5 \\
Kinyarwanda & 6 & 6 (100\%) & 1.00 & 5 \\
Swahili & 6 & 6 (100\%) & -- & 6 \\
Twi & 6 & 5 (83.3\%) & .67 & 2 \\
Yoruba & 6 & 2 (33.3\%) & .00 & 2 \\
\midrule
All & 36 & 29 (80.6\%) & .59 & 23 \\
\bottomrule
\end{tabular}
\caption{Existing native-review agreement for the 36 finalized decomposition templates. The statistics document review coverage and disagreement; they do not establish equal pragmatic force across languages.}
\label{tab:humanreview}
\end{table}

\section{Twi Paraphrase Boundary}

The Twi paraphrase stress test is a boundary condition, not a positive robustness result. For Qwen3-8B, all 48 paraphrase rows are parseable and jointly parseable with the original endorsed T2 condition. Target selection drops from 70.8\% under the original wording to 4.2\% under the paraphrase ($-66.7$\pp). Several other checkpoints have too few jointly parseable rows for a stable comparison. This failure is one reason the paper does not claim native prompt-family invariance.

\section{Follow-up Estimates and Chronology}

The notebook's internal phase labels are provenance identifiers from a larger pipeline; skipped numbers do not represent undisclosed experimental conditions. Table~\ref{tab:chronology} uses scientific stage names and records whether each generation stage was frozen before its own outputs. The final reviewer-resolution stage is analysis-only.

\begin{table*}[h]
\centering
\small
\begin{tabular}{p{2.6cm}p{2.7cm}p{3.0cm}p{5.4cm}}
\toprule
Analysis & Status & Estimate & Interpretation \\
\midrule
Native decomposition & Frozen before own generation & Any-turn E--M+V +29.26\pp [26.68,31.76] & Original composite prompt-family contrast; not isolated assertion. \\
Uncertainty moderation & Post-hoc & Higher T1 margin predicts larger E--M+V difference & Exploratory moderation only. \\
Persistence W1 & Frozen follow-up & +25.94\pp [23.30,28.65] & Within-context carry-over after explicit endorsement stops. \\
Matched truth DID & Frozen follow-up & $-1.10$\pp [$-3.49$,1.37] & Little evidence the increment is larger for false than true targets. \\
All non-gold targets & Frozen follow-up & +28.81\pp [26.82,30.78] & Weakens designated-distractor artifact. \\
Native wording replication & Frozen before own generation & +22.54\pp [20.60,24.41] & Three pre-existing validated standalone wordings. \\
Factorial identification & Precommitted before later robustness generation & Assertive +30.38\pp [28.38,32.35] & English control layer; target mention fixed; three evaluated families. \\
Existing-data reconciliation & Post-hoc; no generation / no new review & Native immediate +19.01\pp; M+V baseline gap +2.43\pp [$-0.02$,4.92] & Aligns metrics, quantifies cross-layer gap, family spread, and review agreement. \\
\bottomrule
\end{tabular}
\caption{Chronology and claim boundaries. ``Frozen'' means fixed before that stage's own generation; it does not imply preregistration of the original project.}
\label{tab:chronology}
\end{table*}

\end{document}